\documentclass[sigplan,10pt]{acmart}
\setcopyright{none}
\usepackage{booktabs}   %% For formal tables:
\usepackage{subcaption} %% For complex figures with subfigures/subcaptions
\newcommand{\stitle}[1]{\vspace{0.5ex} \noindent{{\bf #1}}}

\usepackage{xspace}
\usepackage{multirow}
\usepackage{algorithm}
\usepackage{algorithmicx}
\usepackage{algpseudocode}
\usepackage{enumitem}
\usepackage{bbm}
\setlist[itemize]{leftmargin=*}
\setlist[enumerate]{leftmargin=*}
\renewcommand\footnotetextcopyrightpermission[1]{}

\begin{document}
\title{Revisiting Service Level Objectives and System Level Metrics in Large Language Model Serving}

\author{Zhibin Wang\textsuperscript{1}, Shipeng Li\textsuperscript{1}, Yuhang Zhou\textsuperscript{1}, Xue Li\textsuperscript{2}, Zhonghui Zhang\textsuperscript{1}, Nguyen Cam-Tu\textsuperscript{1}, Rong Gu\textsuperscript{1}, Chen Tian\textsuperscript{1}, Guihai Chen\textsuperscript{1}, Sheng Zhong\textsuperscript{1}}
\affiliation{
  \textsuperscript{1}State Key Laboratory for Novel Software Technology, Nanjing University
  \country{}
  \\
  \textsuperscript{2}Alibaba Group
  \country{}
}

\begin{abstract}
  User experience is a critical factor Large Language Model (LLM) serving systems must consider, where service level objectives (SLOs) considering the experience of individual requests and system level metrics (SLMs) considering the overall system performance are two key performance measures. However, we observe two notable issues in existing metrics: 1) manually delaying the delivery of some tokens can improve SLOs, and 2) actively abandoning requests that do not meet SLOs can improve SLMs, both of which are counterintuitive.

  In this paper, we revisit SLOs and SLMs in LLM serving, and propose a new SLO that aligns with user experience.
  Based on the SLO, we propose a comprehensive metric framework called smooth goodput, which integrates SLOs and SLMs to reflect the nature of user experience in LLM serving. Through this unified framework, we reassess the performance of different LLM serving systems under multiple workloads. Evaluation results show that our metric framework provides a more comprehensive view of token delivery and request processing, and effectively captures the optimal point of user experience and system performance with different serving strategies.

\end{abstract}

\maketitle
\pagestyle{plain}

\section{Introduction}
\label{sec:introduction}

\begin{figure}[t]
    \centering
    \includegraphics[width=1\linewidth]{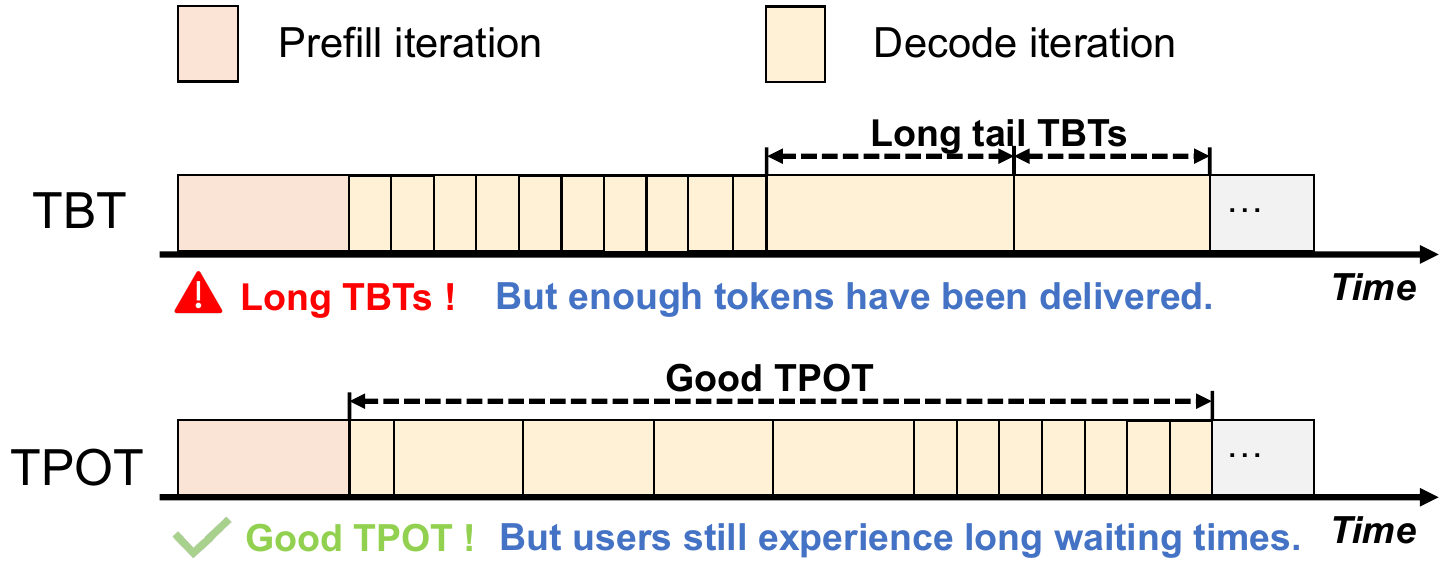}
    \caption{Examples illustrating the limitations of TBT and TPOT.}
    \label{fig:tbt_tpot}
\end{figure}
Given the outstanding performance of large language models (LLMs)~\cite{touvron2023llamaopenefficientfoundation,wake2024yilightningtechnicalreport,qwen2025qwen25technicalreport,google2024gemini,glm2024chatglm,xai2024grok} in various tasks, such as chatbots~\cite{openai2024chatgpt,zheng2024judging,montagna2023data} and virtual assistants~\cite{vu2024gptvoicetasker,dong2023towards}, it has become prevalent for service providers to deploy LLMs to deliver services to users.
With the growing demand for LLM services, the performance of LLM serving systems has garnered significant attention.
Initially, the primary objective of LLM serving systems was to maximize throughput\cite{yu2022orca, kwon2023efficient,cheng2024enablingefficientbatchserving,zhang2024topologyawarepreemptiveschedulingcolocated,zhu2024nanoflowoptimallargelanguage}, significantly improving resource utilization and reducing costs.
However, these optimizations often result in higher request latency since the system handles higher QPS (Queries Per Second), thereby compromising the real-time interactive experience between users and systems~\cite{openai2024chatgpt,deepseekai2025deepseekr1incentivizingreasoningcapability,openai2024openaio1card,vu2024gptvoicetasker,dong2023towards,ong2025routellmlearningroutellms}.

Recently, service providers have begun to prioritize user experience in LLM serving systems, leading to the introduction of service level objectives (SLOs) and system level metrics (SLMs) to evaluate the performance of LLM serving systems~\cite{patel2023splitwise, agrawal2024taming,zhong2024distserve,cheng2025scootsloorientedperformancetuning,patke2024queueneedresolvingheadofline}. Briefly, SLOs are defined as the constraints on the performance of each request, e.g., the latency of the output tokens, while SLMs are defined as the performance of the entire system, e.g., the goodput, which is the throughput of the requests that meet the SLOs.

Specially, to evaluate user experience in LLM serving systems, service level objectives (SLOs) that measure the performance on aligning user experience have been used to in LLM serving systems~\cite{patel2023splitwise,agrawal2024taming,cheng2025scootsloorientedperformancetuning,patke2024queueneedresolvingheadofline}, such as time-to-first-token (TTFT), time-between-tokens (TBT), and time-per-output-token (TPOT).
Due to the autoregressive nature of LLMs, the generation of first token (i.e. prefill phase)~\cite{vaswani2023attentionneed,zhong2024distserve} is more costly than the generation of subsequent tokens (i.e. decode phase) since the prefill phase requires the model to process the entire prompt, while the decode phase only requires processing the current token.
Considering both the user experience and the characteristics of prefill phase, TTFT is introduced to measure the time required to generate the first token, which may be significantly larger than TBT or TPOT.
TPOT measures the average time between tokens in a request, while it is too loose to reflect the user experience, as a long stall in the middle of the request can be averaged out by short intervals between other tokens, which actually degrades the user experience. Therefore,Sarathi-Serve~\cite{agrawal2024taming} introduces the TBT metric to constrain the time interval between two consecutive tokens.

To further evaluate the performance of LLM serving systems ensuring the SLOs, system level metrics that measure the performance of each request of the system such as SLO attainment and goodput are proposed~\cite{zhong2024distserve,agrawal2024taming}.
The SLO attainment measures the proportion of requests that meet the SLOs, which can be viewed as the constraint of the serving system, while the goodput measures the number of completed requests that meet the SLOs per second, which can be viewed as the performance of the serving system.
We also notice that various systems and optimization strategies have been proposed to improve the system level metrics under the SLOs~\cite{patel2023splitwise, agrawal2024taming,zhong2024distserve,cheng2025scootsloorientedperformancetuning,patke2024queueneedresolvingheadofline}.

However, we observe that \textit{these metrics (i.e. SLOs and SLMs) fail to capture the nature of user experience}.
Actually, real-time LLM service is a rapidly interactive activity, just like web browsing~\cite{weinreich2008not,skadberg2004visitors}. Users do not perceive them as a sequence of individual tokens (modeled as TBT), but as a continuous stream of information where users need to process the earlier tokens before the next token arrives.
The evaluation bias caused by ignoring the inherent nature of user experience in streaming LLM serving can even lead optimization efforts based on these metrics to develop in a suboptimal direction. We identify several limitations in the existing metrics as follows:

\stitle{Limitation 1: In existing SLOs, TBT is too tight for overall user experience while TPOT and E2E latency are too loose.}
TBT measures the time interval between each token within a request, while TPOT reflects the average interval. As indicated in~\cite{egger2012waiting}, user experience in streaming services is influenced by waiting times without information to process.
As shown in Figure \ref{fig:tbt_tpot}, if users have enough information to process, occasional stalls (i.e., high TBT) may not degrade the experience.
For example, if a system delivers 10 tokens in the first second, then stalls for 1 second, users reading at 4 tokens per second will still have a good experience, although the TBT is up to 1 second.
Conversely, if only 2 tokens are delivered before a 1-second stall, users will suffer from the waiting time, although the TPOT is only 0.1s.
In other words, the cost of high latency iterations is shared with \textit{previous} iterations.

\stitle{Counterintuitive Example 1:} To show the limitations of existing SLOs, consider the trick that will manually delay the delivery of all tokens to the TBT threshold. For example, if the TBT threshold is 200ms, the system will not deliver the current token until 200ms has passed after the delivery of last token.
In the first example of Figure~\ref{fig:tbt_tpot}, assuming the TBT threshold is 200ms, the 10 tokens generated rapidly within the first second will not be delivered to the user when generated. Instead, they are manually delayed until the interval reaches 200ms, thereby providing sufficient buffer time for the slower token generation later, significantly improving TBTs.
However, this trick actually delays the delivery time of all tokens. For the user, the arrival time of each token is no earlier than it would be under the strategy of immediate delivery upon generation. This obviously results in a worse experience, yet it achieves better performance under the existing SLO framework, which is counterintuitive.

\stitle{Limitation 2: In existing SLMs, Goodput and SLO attainment are not able to reflect the benefits of requests that exceed the SLO.}
Goodput is a system level metric that can reflect the number of completed request that meet the SLOs per second, {while SLO attainment reflects the proportion of requests that meet the SLOs.}
However, existing metrics definitions ignore the contribution of requests that are missed.
Therefore, the optimal strategy seems to be to give up or reject the requests that have already missed the SLOs, which is not a good choice for users obviously.
We argue that the requests that missed the SLO requirements are still valuable, and the benefits of all the requests should be carefully considered.

\stitle{Counterintuitive Example 2:} To show the limitations of existing system level metrics, consider the trick that will actively abandon the requests that have already missed the SLOs.
Since the goodput of these request is always 0, abandoning them will result in no loss of goodput, but will improve the throughput of the remaining running requests by reducing the contention for resources, leading to a higher goodput.
In LLM serving, however, only considering these metrics is an unacceptable outcome for users.
While latency undoubtedly degrades the user experience, abandoning a request altogether poses an even greater threat, which is not considered in the existing system level metrics.

\begin{figure}[t]
    \centering
    \includegraphics[width=1\linewidth]{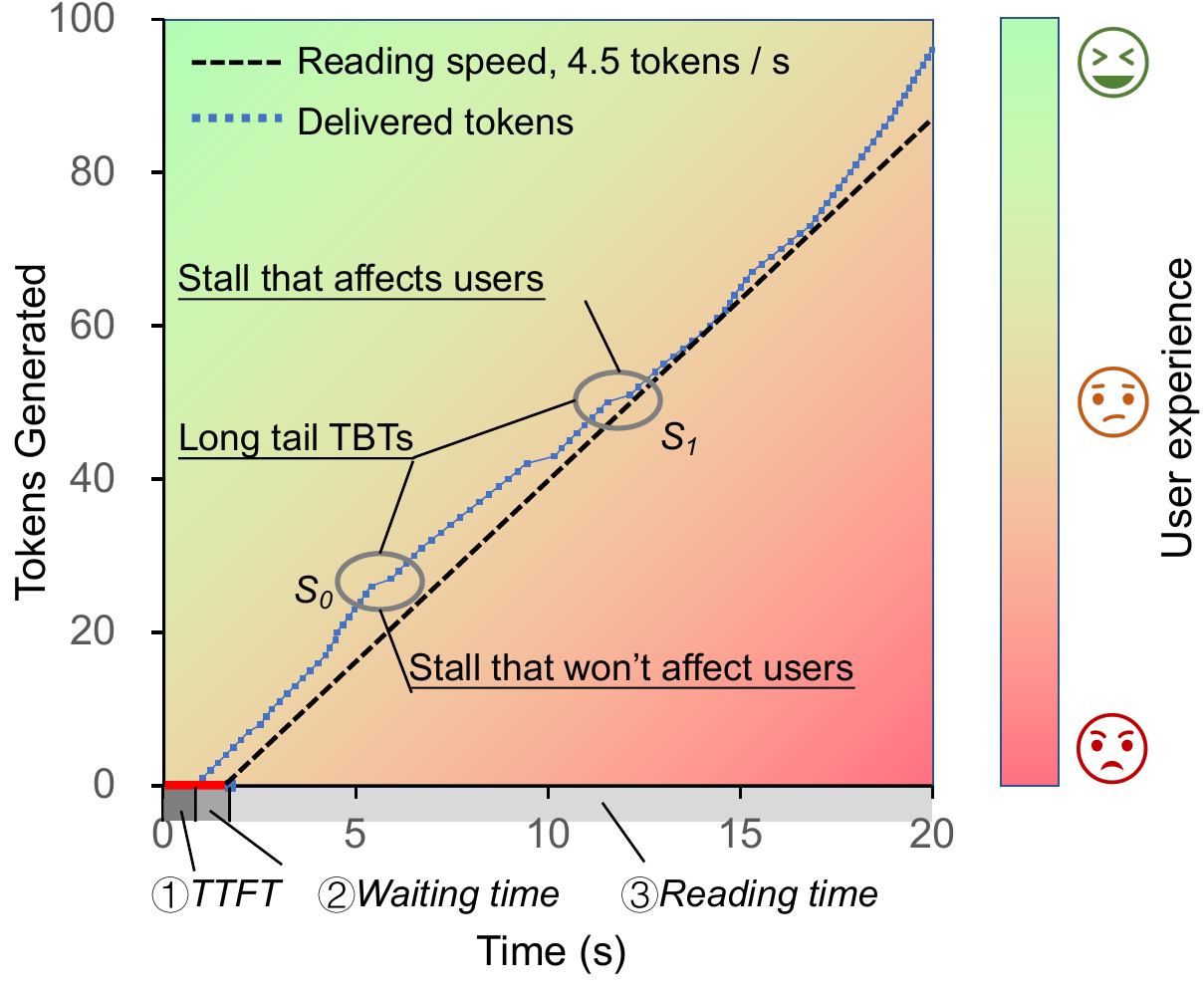}
    \caption{Token generation timeline in LLM serving systems and its impact on user experience. The red area indicates affected user experience, while the green area indicates good user experience. The blue line represents the token generation timeline and each dot represents one token.
        The total timeline can be devided into three parts: \textcircled{1} the time to receive the first token (TTFT), \textcircled{2} the time when the user is waiting for the next token, and \textcircled{3} the time when the user is consuming the delivered information.
    }
    \label{fig:example of output}
\end{figure}
Combining the SLOs and SLMs, Figure~\ref{fig:example of output} illustrates how the streaming token delivery affects user experience and how the contribution to smooth goodput of a request is determined by the token delivery timeline.

In the timeline, at the beginning, the user has to wait for the first token. Then, subsequent tokens are delivered quickly, causing no decrease in user experience.
The slow delivery of tokens in the middle of the timeline (from t=10 to t=15) seriously degrades the user experience, as the user has to wait for the next token without any information to process. Since the affect has already occurred, fast delivery of later tokens cannot compensate for the waiting time. Therefore, the contribution of the request to the smooth goodput is determined by the total waiting time, denoted by the reference line tangent to the delivery timeline, with its slope representing the reading speed (4.5tokens/s in this case). The color region where the reference line is located indicates the actual contribution of this request to the smooth goodput—green (top-left region) represents higher contribution, while red (bottom-right region) indicates poorer contribution.
Meanwhile, from the delivery timeline we can see that occasional stalls cause long tail TBTs, but will not affect the user experience as long as the user has enough tokens to read. Specifically, the user may not even notice the stall $S_0$ although the TBT is large as they are reading the delivered information before $S_0$.
On the other hand, the user suffers from the waiting time at $S_1$, since the delivered tokens before $S_1$ have been consumed.

In this paper, we revisit system level metrics and SLOs in LLM serving systems, highlighting their limitations in modeling user experience. To address this, we propose a redesigned SLO metric that defines token deadlines relative to the request commitment rather than the previous token. Building on this, we introduce smooth goodput, a novel performance metric that balances the benefits of token generation against the penalty for user idle time during token delivery. Based on this unified framework, we re-evaluate the performance of different LLM serving systems under multiple workloads, aiming to help unify the development direction of research on LLM serving focused on user experience optimization.

Our contributions include:

\begin{itemize}
    \item We revisit the existing SLOs and system level metrics in LLM serving systems, and identify their limitations in modeling user experience through two counterintuitive examples.
    \item We propose a new SLO metric to define the deadlines of each token relative to the commitment of a request, rather than relative to the previous token, which aligns more closely with how users process information.
    \item We introduce the smooth goodput metric based on the new SLO metric to evaluate the performance of LLM serving systems, which considers the benefit of all generated tokens based on their contribution to the user experience and throughput.
    \item We conduct extensive experiments with our framework to evaluate the performance of different LLM serving systems and address the counterintuitive issues in user experience and system performance caused by the existing SLOs and SLMs. We show our framework provides a more comprehensive view of token delivery and request processing.
\end{itemize}

\section{Background}
\label{sec:background}

In this section, we provide background on LLM services, including the principles of autoregressive inference, methods for evaluating user experience, and specific metrics.

\subsection{LLM Inference}

LLMs process autoregressive inference to generate output tokens based on input prompts and the previous generated tokens.
Specifically, a prompt of length $k$ can be represented as a token sequence ($t_1, t_2, ..., t_k$).
The output generated by the LLM is also a token sequence of length $n$, denoted as ($t_{k+1}, t_{k+2}, ..., t_{k+n}$).
The entire inference process consists of $n$ iterations, where each iteration generates a token, as known as the autoregressive process.
Due to the autoregressive nature of LLMs, the generation of each subsequent token requires the KV state of all previous tokens, so the previous KV states are cached for reuse, known as the KV cache.

Based on the characteristics of computation and memory access, these iterations can be divided into two phases: \textit{prefill} and \textit{decode}.
As shown in Fig.~\ref*{fig:metrics}, in the prefill phase, the LLM processes the entire prompt within a single iteration ($A^P$) which is longer than the decode phase. In the prefill phase, the LLM compute the KV state for each token. Highly parallelizable matrix multiplications are performed, making the prefill phase compute-bound~\cite{agrawal2023sarathiefficientllminference}.
The following decode phase contains the subsequent iterations ($A^D_1, A^D_2, ..., A^D_n$), ending with the generation of the EOS (End of Sequence) token $A^D_{EOS}$.
In a iteration of decode phase, the prompt and the tokens generated in previous iterations are concatenated as the input, with their KV state cached in memory, requiring no recomputation but more memory access.
Low aritmetic intensity makes decode phase memory-bound, and the computation is dominated by memory access~\cite{agrawal2023sarathiefficientllminference}.
To mitigate interference between the two computationally distinct phases, recent works~\cite{zhong2024distserve,patel2023splitwise,hu2024inferenceinterferencedisaggregatellm} proposes decoupling the prefill and decode phases by deploying them on separate instances. While this incurs additional communication overhead, it creates opportunities for phase-specific resource customization, thereby significantly improving the efficiency of each phase.

\subsection{LLM Serving}
\stitle{LLM serving task.} In practice, LLMs are deployed as services to provide users with inference capabilities. In terms of the interaction mode with users, these services can be broadly categorized into two types of modes: online serving and offline serving.

\textit{Online LLM serving} are often designed to provide real-time services to users~\cite{sun2024llumnixdynamicschedulinglarge,agrawal2024taming,kossmann2025gpuhalfemptyhalffullpractical,wu2024fastdistributedinferenceserving,cheng2025scootsloorientedperformancetuning,jiang2024neosavinggpumemory}, which is a rapidly interactive activity like web browsing~\cite{weinreich2008not,skadberg2004visitors}. When interacting with LLMs, users expect the system to respond quickly and provide instant feedback, and consume information in a continuous stream.
During this continuous stream of information, users expect not to suffer from long waiting times, neither for the first response token nor for subsequent tokens~\cite{egger2012waiting}.
Existing works~\cite{brysbaert2019many} has studied the speed of reading and processing text. The average reading speed of an adult is about 3-4 words per second. Based on the granularity of tokenization in different languages, we can roughly estimate the token generation speed target.

\textit{Offline LLM serving} provides non-streaming service, where user experience is not as stringent as in online scenarios~\cite{qiao2024conserveharvestinggpuslowlatency,wang2025echoefficientcoschedulinghybrid,zhao2024blendserveoptimizingofflineinference}. Users generally focus on the end-to-end metrics of batched offline tasks and typically do not have specific requirements for streaming-specific metrics like TBT and TPOT. We mainly focus on the online LLM serving in this paper, where the user experience is critical.

\begin{figure}[t]
    \centering
    \includegraphics[width=1\linewidth]{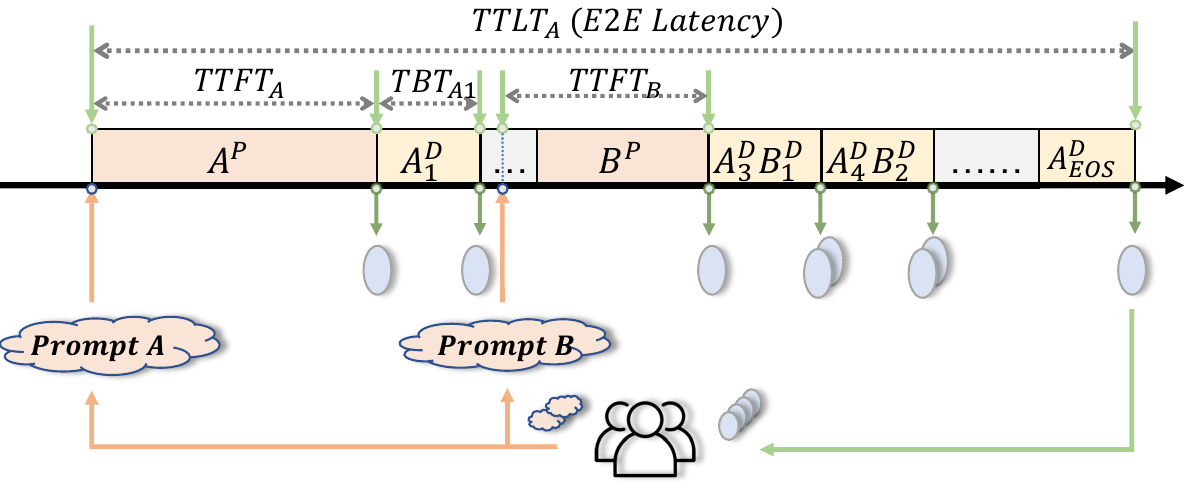}
    \caption{Existing SLOs of LLM Serving. Note that this figure ignores the difference between token generation from the LLM and its delivery to users.}
    \label{fig:metrics}
\end{figure}

\stitle{LLM serving metrics.} 
To ensure a consistently high-quality user experience and efficient resource utilization, providers monitor and evaluate the performance of LLM serving systems using various metrics.
Based on the considerations, these metrics used to evaluate the performance of LLM serving can be divided into two main groups:

\textit{Service Level Objectives (SLOs, in Section~\ref{sec:revisit_slo})} are the metrics that directly represent the token delivery state of a request. These include response time metrics such as TTFT, TBT, and TPOT, which help assess whether the request can deliver responses within the expected time frame.
By defining SLOs, providers establish observable thresholds to proactively assess compliance with user expectations, which is critical in dynamic online serving environments.
As shown in Figure \ref{fig:metrics}, To ensure that users receive the expected quality of service, these SLOs focus on aspects such as TTFT (Time-to-First-Token), TBT (Time-between-Tokens), TPOT (Time-per-Output-Token) and so on. These metrics are crucial for evaluating the responsiveness of the system and ensuring that users receive timely feedback during their interactions with the LLM.

\textit{System Level Metrics (SLMs, in Section~\ref{sec:revisit_goodput})} are the metrics that assess the performance and operational efficiency of the infrastructure under the constraints of SLOs. This category includes system throughput, resource utilization, and user experience. These metrics include evaluations of SLO attainment~\cite{liu2024andes}, goodput~\cite{zhong2024distserve}, capacity~\cite{agrawal2024taming} and so on. Such metrics are key to reflecting performance effectiveness and ensuring that the system can meet both cost and experience demands.

By combining these SLOs and SLMs, service providers can obtain a comprehensive view of performance. Therefore, revisiting these metrics is essential for understanding the performance of LLM serving systems and their impact on user experience.

\section{Revisiting the SLOs of Each Request}
\label{sec:revisit_slo}

In this section, we focus on the user experience of each request, i.e., the SLOs.
We first introduce the framework of SLOs, which can be customized to represent various requirements proposed in different workloads, as well as revisiting the existing SLOs in recent works on LLM serving.
Then, we demonstrate a output delay trick that can be used to improve the SLO attainment on existing SLOs, which is counterintuitive.
Finally, we propose a new SLO that is more aligned with user experience, focusing on the relationship between the information processing of the user and the delivery of information by the service.

\subsection{SLO Framework}

Recall the progress of the generation process in LLM serving, the output of a request is form of a sequence of tokens, where the delivery time of each token is crucial for user experience.
Thereby, we introduce a unified framework of SLOs that can be customized to represent the various requirements proposed in different workloads.
No matter in which SLO, the request level target is to measure whether the token generation time meets the threshold, so we define the unified SLO as the deadline of each token.

\stitle{Framework definition.} For a request $r$ with output length as $n$, $i$ is the index of the output tokens, which starts from 1 and ends with $n$.
We define the deadline of the $i$-th output token of a request as $d_{i}$, while $t_{i}$ is the actual generation time of the $i$-th output token.
Therefore, the SLO constraints can be formulated as:
\begin{equation}
    \forall i, t_{i} \leq d_{i}.
\end{equation}

\stitle{Customization of existing SLOs.} The framework can be customized to represent the various requirements proposed in different works by adjusting the deadline of each token. The details customization of existing SLOs are following:
\begin{itemize}
    \item \textbf{TTFT (Time-to-First-Token) and TBT (Time-between-Tokens):} TTFT reflects the time taken for the generation of the first output token while TBT represents the fine-grained time interval between two adjacent tokens of a request. They delve into each token generation process. Thus, with our framework, we can define the SLO on TTFT and TBT as:
          \begin{equation}
              d_{i} = \left\{
              \begin{array}{ll}
                  TTFT_\theta,          & i=1, \\
                  t_{i-1} + TBT_\theta, & i>1,
              \end{array}
              \right.
          \end{equation}
          where $TTFT_\theta$ and $TBT_\theta$ are the thresholds of TTFT and TBT, respectively~\cite{agrawal2023sarathiefficientllminference,agrawal2024taming}. The SLO is met if the generation time of the first token is less than $TTFT_\theta$, and the generation time interval between adjacent tokens is less than $TBT_\theta$.
          Note that the deadline of the $i$-th token is determined by the generation time of the previous token, which, as we will show, is not aligned with user experience.
    \item \textbf{TTFT and TPOT (Time-per-Output-Token):} TPOT reflects the average time taken to generate a token (excluding the first token). With our framework, we can define the SLO on TTFT and TPOT as:
          \begin{equation}
              d_{i} = \left\{
              \begin{array}{ll}
                  TTFT_\theta,                 & i=1, \\
                  t_{1} + (n-1) * TPOT_\theta, & i>1,
              \end{array}
              \right.
          \end{equation}
          where $TPOT_\theta$ is the threshold of TPOT~\cite{patel2023splitwise,zhong2024distserve,qin2024mooncakekvcachecentricdisaggregatedarchitecture}. The SLO is met if the generation time of the first token is less than $TTFT_\theta$, and the generation time of all tokens in the request is less than the sum of $t_1$ and the product of the maximum length and the tpot threshold, i.e., $t_{1} + (n-1) * TPOT_\theta$.
          In other words, only the first and last tokens are subject to timing constraints.
    \item \textbf{End-to-end latency:} E2E latency reflects the total time taken for a request (or a batch of requests) from commited by users to when it completed. With our framework, we can define the SLO on E2E latency as:
          \begin{equation}
              d_{i} = E2E_\theta,
          \end{equation}
          where $E2E_\theta$ is the threshold of end-to-end latency.
          Obviously, if the last token is generated before the end-to-end latency, the request meets the SLO. As aforementioned, the end-to-end latency and TPOT is a very loose constraint, which is not aligned with user experience all the time.
\end{itemize}

Obviously, the TPOT can be seen as a generalization of E2E latency, therefore, we will not discuss the E2E latency in the following sections.

\subsection{Counterintuitive Example of Existing SLOs}
We start with a counterintuitive example of existing SLOs, which is the output delay trick.

\stitle{Output delay trick.}
Output delay is a tactic where tokens are released until the TBT deadline is reached, rather than immediately upon generation.
Specifically, delaying the delivery of the $i$-th token can give more loose constraints on the generation time of the $i+1$-th token under the TBT SLO. The generation time of the $i+1$-th token can be loosed from $TBT_\theta$ to $TBT_\theta + t_{delay}$, where $t_{delay}$ is the delay time of the $i$-th token.
Implementing output delay can be easily done by adding an intermediate buffer layer between the inference engine and the client.

However, this trick is irational in the context of user experience, delaying the delivery of the tokens actually affects the user experience as users see the tokens later than they are generated.
Essentially, it is because the premature delivery of tokens inadvertently imposes additional latency constraints on the subsequent tokens.
Thus, there is an urgent need to devise a novel SLO that not only protects the user experience but also refrains from penalizing the early delivery of tokens.

Subsequently, we will introduce a optimization strategy called \textit{chunked prefills} misleading by the TBT SLO, and then we will propose a naive imitation strategy called \textit{decode prepone} that can achieve a comparable effect to chunked prefills on TBT by simply scheduling and output delay tricks.

\begin{figure}[t]
    \centering
    \includegraphics[width=1\linewidth]{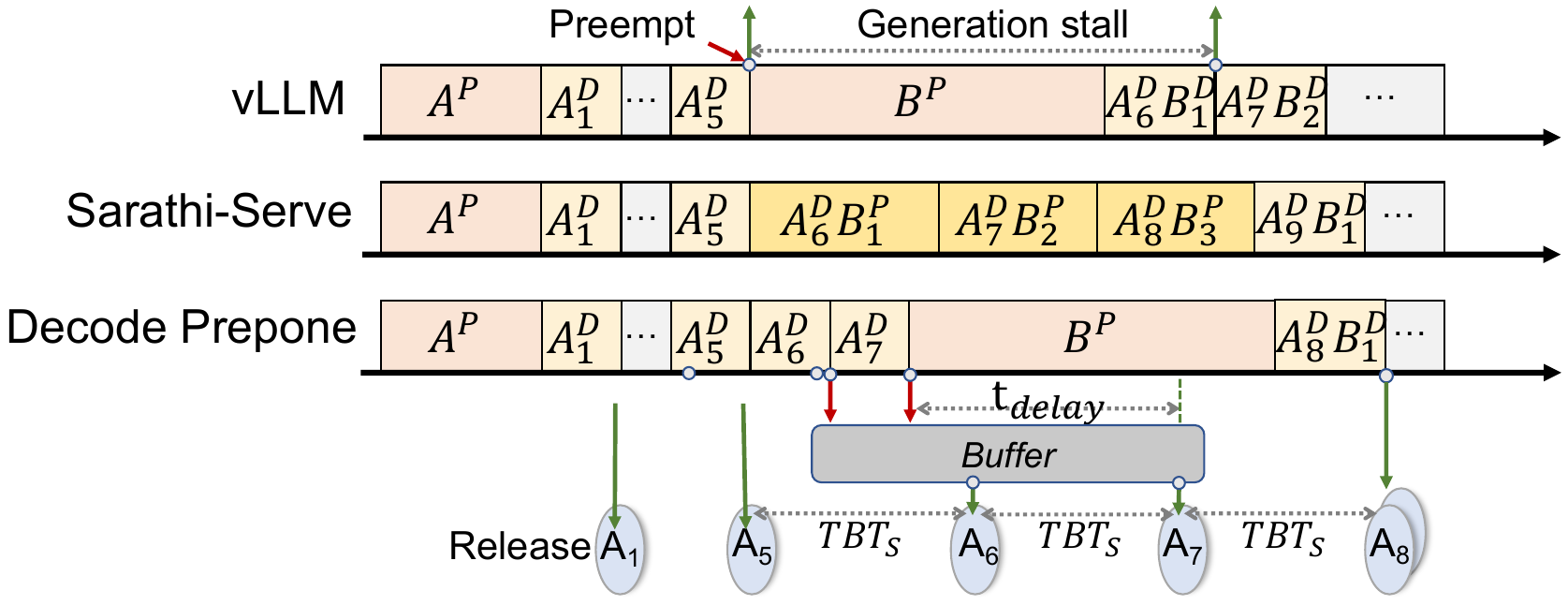}
    \caption{An illustration of iteration scheduling strategies.}
    \label{fig:prepone}
\end{figure}

\stitle{Chunked prefills.} As shown in Figure~\ref{fig:prepone}, we illustrate the generation process of two requests A and B under sarathi-serve framework~\cite{agrawal2024taming} which is a typical chunked prefills strategy.
Due to the prefill-prioritizing principle for improving throughput in vLLM, the decode phase of the following tokens for request A will be stalled until the prefill phase of request B is finished, which results in a generation stall, i.e., a large TBT between $A_5^D$ and $A_6^D$.
Therefore, Sarathi-Serve splits the prefill phase of B into multiple chunks ($B_1^P$, $B_2^P$, $B_3^P$) and fuses them with the decode phases of request A in the same batch.
Specifically, one prefill chunk of request B will attach decoding one token of request A, like $A_6^DB_1^P$, $A_7^DB_2^P$ and $A_8^DB_3^P$.
Assuming the prefill stage of B is split into $n_c$ chunks, the stall time of A is approximately reduced to about $\frac{1}{n_c}$ of the original.
By this way, the stall time is smoothed, resulting in a smaller TBT.
However, we observe that the absolute latency from decode tokens of request B ($B_1^D, B_2^D....$) will not benefit from the optimization.
Further, our concern arises that this slicing approach, by introducing frequent assessments of the KV cache, may inadvertently lead to an increase in overall latency rather than a decrease.

To summarize, the chunked-prefills smooths the TBT by slicing the prefill phase and fusing them with the decode phases of other requests.
This provides an insight that instead of slicing, can we manually schedule the prefill and decode phases and achieve better performance?

\stitle{Decode prepone.}
We propose a naive imitation strategy, called \textit{decode prepone}, which can achieve a comparable effect to chunked prefills on TBT by simply scheduling without slicing.
As shown in Figure~\ref{fig:prepone}, specifically, the next $n$ decode tokens for request A ($A_6^D$ and $A_7^D$) are preponed to be generated before the prefill of request B starts.
Meanwhile, instead of directly outputting these tokens of request A, which can result in large TBT between $n$-th token ($A_7^D$) to $n+1$-th token ($A_8^D$), we smoothly output these tokens during the prefill phase of request B.

To achieve smooth output, we take an intuitive approach by assigning a $t_{delay}$ to the output timing of each preponed token.
As shown in Figure~\ref*{fig:prepone}, even though $A_6^D$ and $A_7^D$ have completed their decoding, they are scheduled to be released sequentially after the $t_{delay}$ interval, while ensuring their output time will not exceed the completion time of B's prefill phase.
This strategy smooths the overall output flow while maintaining overall latency and mitigating excessive TBT concerns.
Besides, it can also be adopted to trade off TTFT for TBT or TPOT by delaying the delivery of the first token.

The delay output trick is reflected the irrationality of existing SLOs, while decode prepone further demonstrates that the optimization targeting for these irrational SLOs is not necessarily beneficial to user experience.

\subsection{A New SLO Defination}
\stitle{Intuition.}
In fact, users do not frequently notice the lag of the last word during the generation process. We argue that generation stalls are not necessarily harmful to user experience, as long as the delivery of tokens is aligned with the user's reading speed.
Given the limitations of TBT in setting the time interval between adjacent tokens, we shift the focus of the SLO to the actual user experience.
For instance, we can set the constraint of each request according to the response delay that users can tolerate and the speed of processing output information, such as reading the output of the chatbot, understanding the summary of long text, listening, etc.

\stitle{Definition:} Porting the new SLO to the framework, we have
\begin{equation}
    d_{i} = V \times i,
\end{equation}
where $V$ is the output information processing speed of the user, and $i$ is the index of the output words. $d_i$ constraints the deadline of the $i$-th token, after which the user will perceive a pause in the output stream.

\section{Revisiting the SLMs}
\label{sec:revisit_goodput}

Based on the SLOs defined on the request level, system level metrics have been proposed to measure the performance of the service. In this section, we first revisit the existing system level metrics, including SLO attainment and goodput, and discuss their shortcomings in LLM serving. We then propose a new metric called \textit{smooth goodput} to measure the performance of LLM serving.

\subsection{Formulation of SLMs}
SLOs are only concerned with the user experience at request level.
However, in the system view, the service provider is more concerned about the overall performance of the service.

Before introducing our smooth goodput, we first revisit the existing system level metrics, including SLO attainment and goodput, and discuss their shortcomings in LLM serving.

For a service of LLM, we define the set of requests as $R$, and $|R|$ is the total number of requests. Thus, we have:
\begin{itemize}
    \item \textbf{SLO Attainment:} SLO attainment is used to describe the proportion of requests that meet the SLOs. T SLO attainment of a service can be defined as:
          \begin{equation}\label{eq:slo_attainment}
              \text{SLO attainment} = \frac{\sum_{r\in R} \mathbbm{1}(\forall i,t_i\leq d_i)}{|R|},
          \end{equation}
          where $\mathbbm{1}(\cdot)$ is the indicator function that returns 1 if the request meet the SLO and 0 otherwise.
          Based on it, capacity is defined as the maximum request rate under the constraint of certain SLO attainment.
    \item \textbf{Goodput:} Goodput is defined as the number of completed requests that meet the SLOs per second in a service. It considers the trade-off between the resource utilization and user experience.
          Formally, the goodput of a service is defined as:
          \begin{equation}\label{eq:goodput}
              \text{Goodput} = \frac{\sum_{r\in R} \mathbbm{1}(\forall i,t_i\leq d_i)\cdot n_r}{T},
          \end{equation}
          where $T$ is the time interval of serving the requests in $R$, and $n_r$ is the number of tokens that the request $r$ generates.
\end{itemize}

\subsection{Counterintuitive Behavior of Existing Metrics}

We observe that if a request does not meet the SLOs, its contribution to SLO attainment and goodput is 0.
This approach, when optimizing for the system level metrics, often leads to abandoning requests that cannot meet the SLOs~\cite{qin2024mooncakekvcachecentricdisaggregatedarchitecture}.
Formally, for a request $r$ that has missed the SLO in generation of the $i$-th token, we have:
\begin{equation}
    \mathbbm{1}(\forall i,t_i\leq d_i) = 0,
\end{equation}
which definately contributes 0 to the goodput and SLO attainment.
The goodput-optimal scheduling strategy should kill this request and prioritize the next request that can meet the SLOs, otherwise the resource will be wasted on the request that cannot meet the SLOs.

While latency undoubtedly degrades the user experience, abandoning a request altogether poses an even greater threat. Assuming you are reading the answer to a question, if the system stops generating tokens in the middle, it will be extremely frustrating compared to waiting for a few more seconds to receive the complete answer. Moreover, the users may relaunch the request, as they need the answer, which will further increase the system load and result in a waste of resources and longer waiting time for the users.

Therefore, we argue that the existing goodput and SLO attainment metrics are not suitable to serve as the optimization objectives for LLM serving systems. Actively abandoning requests that do not meet the SLOs is counterintuitive and can lead to a poor user experience. In the next section, we propose a new metric called \textit{smooth goodput} to address this issue.

\subsection{Smooth Goodput}
Given the shortcomings of the existing goodput metric, a new metric must comprehensively consider the contribution of each request, even if it slightly exceeds the SLO requirements. In such cases, users have to wait for the subsequent token to be generated, after they have finished reading all the previously delivered tokens.

\stitle{Streaming service and user experience.}
Unlike models with a single forward inference process, interactive LLM applications are typically deployed as streaming services due to the autoregressive nature of LLMs. Research \cite{egger2012waiting} on web based streaming services has shown that the waiting time of users is a key factor affecting user experience.

Therefore, we introduce the concept of user wait time, namely \emph{user idle latency}, to measure the user experience.
The user idle latency is cumulative duration during which a user is idle and waiting for new tokens to be generated due to the lower generation speed.
Formally, the user idle latency $l$ of a request $r$ is defined as:
\begin{equation}
    l_r = \max_{i=1}^{n}(t_i-d_i),
\end{equation}
where $t_i$ is the time when the $i$-th token is generated, $d_i$ is the deadline time of the $i$-th token delivered to the user, and $n$ is the number of output tokens in the request $r$. In practice, the user idle latency is the total time that the user has to wait for the next token to be generated without any other token left to read during the generation process of the request. User experience will get worse if the user idle latency is large.

\stitle{Definition.}
The smooth goodput is defined as the service benefit per unit of time.
The benefit of a request is defined by two factors: the number of tokens that the request generates and the user idle latency of the request.
Intuitively, token generation timeline represents the service provider's revenue stream, while user idle time results in a loss of user experience. The higher the throughput and the shorter the idle time, the greater the potential profit that the service can generate.
Formally, we have:
\begin{equation}
    \text{benefit}(r) = n_r- \alpha\cdot f(l_r),
\end{equation}
where $n_r$ is the number of tokens that the request $r$ generates, $f(\cdot)$ is a function that maps the user idle latency to the percentage of the benefit that the request can generate, and $\alpha$ is a weight.
For interactive applications with stringent latency requirements, a higher value of $\alpha$ should be chosen to ensure that idle latency is minimized. In practical deployments, the parameters of the benefit function can be calibrated using historical workload data, including request latency metrics and user behaviors (e.g., cancellations and complaints), to better align the service characteristics with the benefit calculation.
\begin{figure*}[t]
    \centering
    \includegraphics[width=0.6\linewidth]{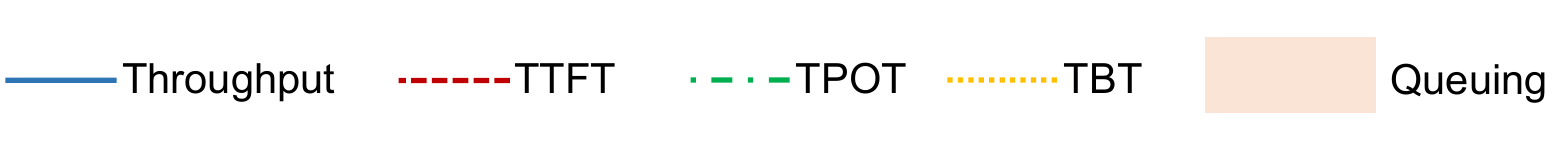}
    \vspace{0.0cm}

    \begin{subfigure}{0.245\textwidth}
        \centering
        \includegraphics[width=0.95\linewidth]{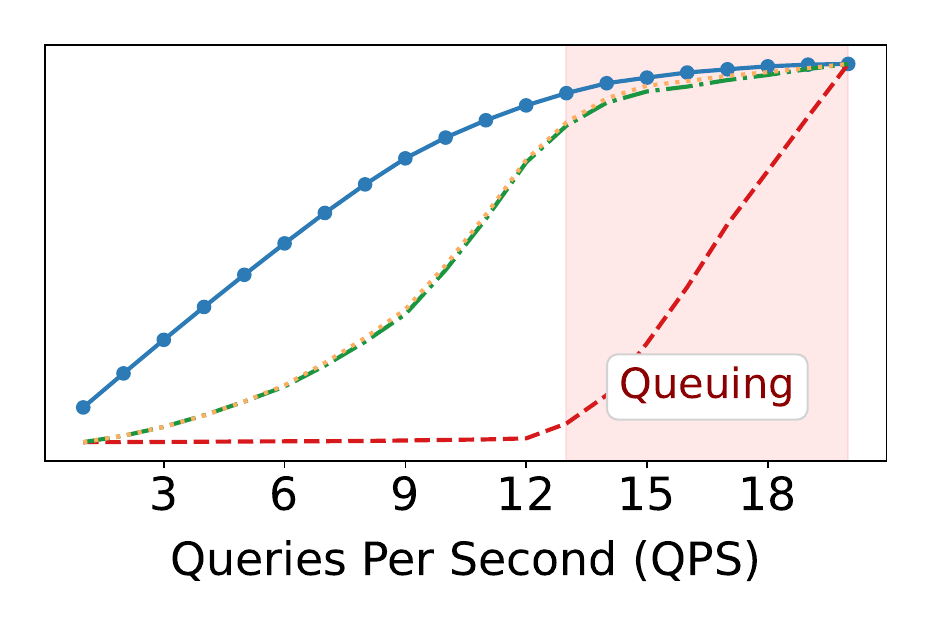}
        \caption{LLaMA-3.1-8B.}
        \label{fig:eval_existing_llama}
    \end{subfigure}
    \hfill
    \begin{subfigure}{0.245\textwidth}
        \centering
        \includegraphics[width=0.95\linewidth]{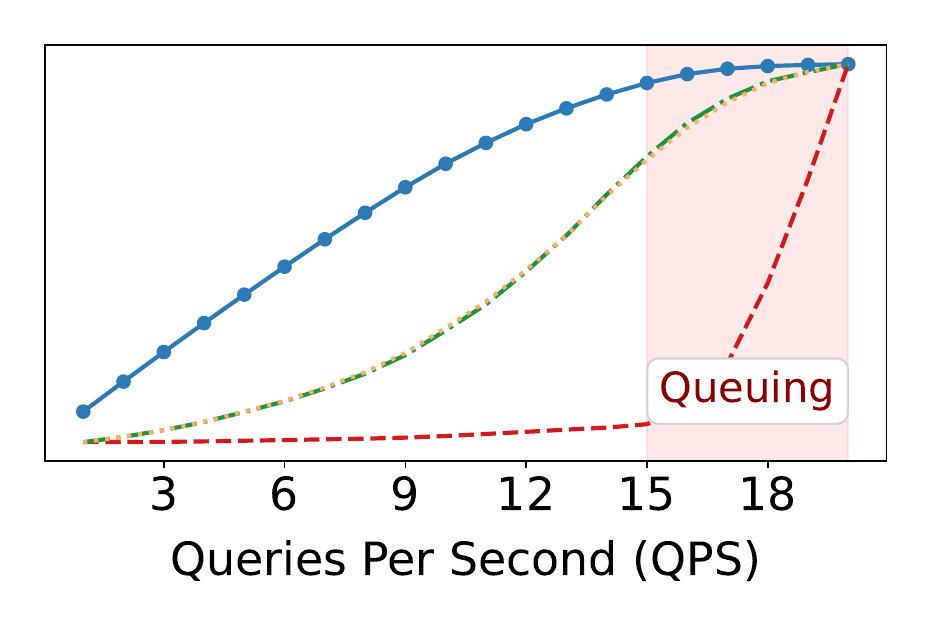}
        \caption{LLaMA-3.1-8B with CP.}
        \label{fig:eval_existing_llama_chunked}
    \end{subfigure}
    \hfill
    \begin{subfigure}{0.245\textwidth}
        \centering
        \includegraphics[width=0.95\linewidth]{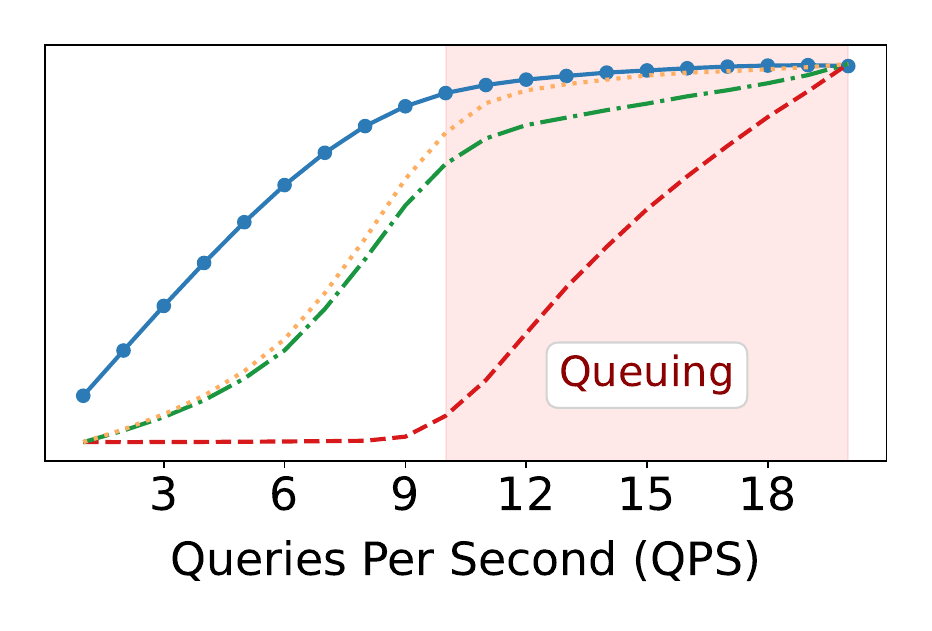}
        \caption{Qwen2.5-14B.}
        \label{fig:eval_existing_qwen}
    \end{subfigure}
    \hfill
    \begin{subfigure}{0.245\textwidth}
        \centering
        \includegraphics[width=0.95\linewidth]{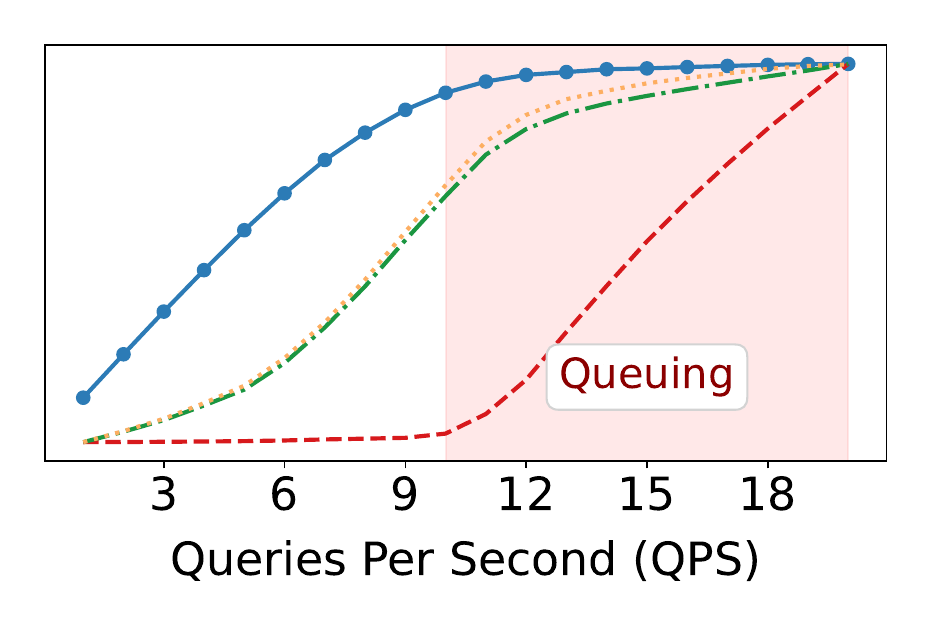}
        \caption{Qwen2.5-14B with CP.}
        \label{fig:eval_existing_qwen_chunked}
    \end{subfigure}
    \caption{Performance of vLLM under different QPS, where CP denotes chunked prefills adoption.}
    \label{fig:eval_existing}
\end{figure*}

\begin{figure*}[t]
    \centering
    \begin{subfigure}{0.245\textwidth}
        \centering
        \includegraphics[width=0.95\linewidth]{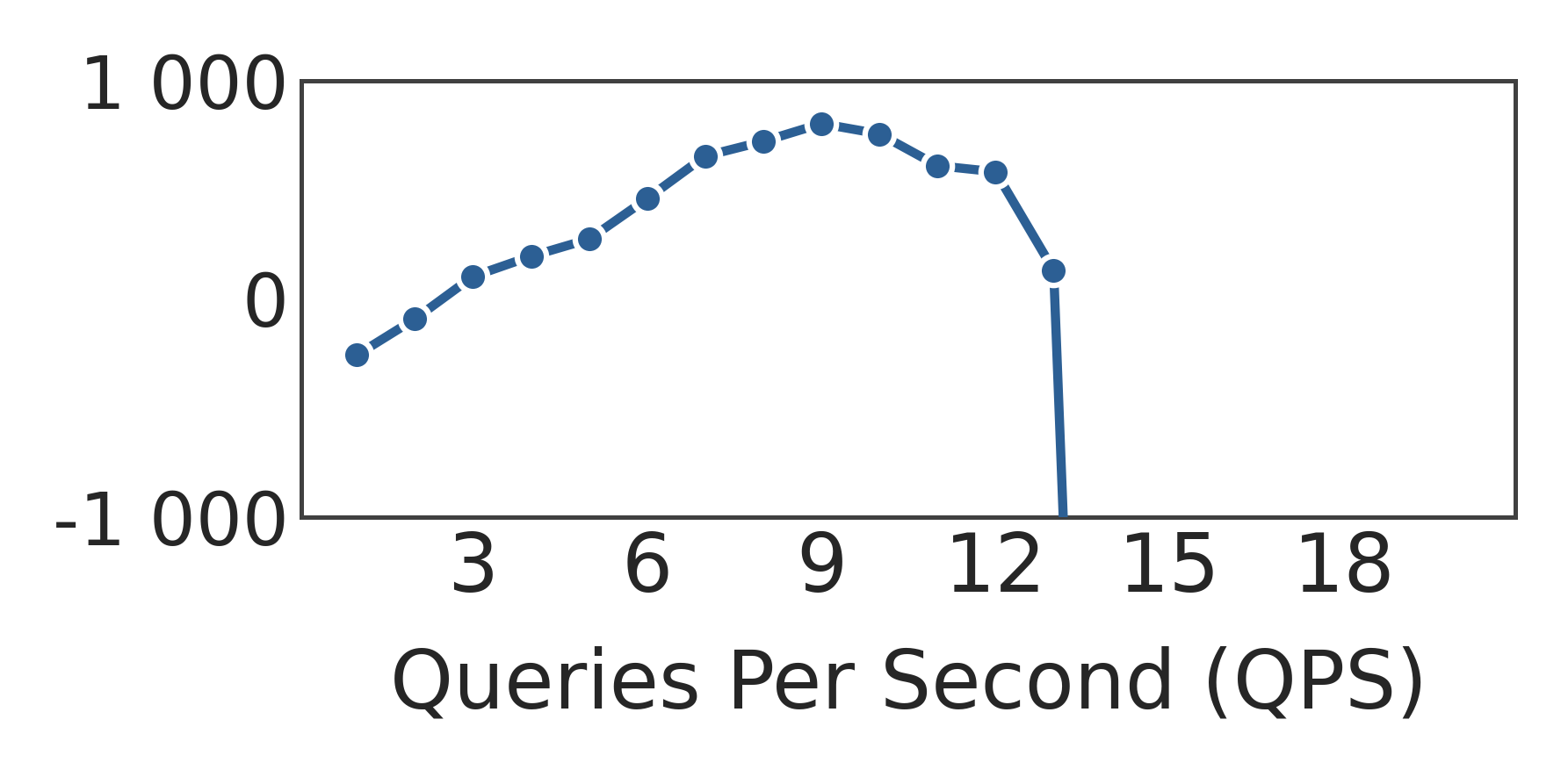}
        \caption{LLaMA-3.1-8B.}
        \label{fig:eval_goodput_llama}
    \end{subfigure}
    \hfill
    \begin{subfigure}{0.245\textwidth}
        \centering
        \includegraphics[width=0.95\linewidth]{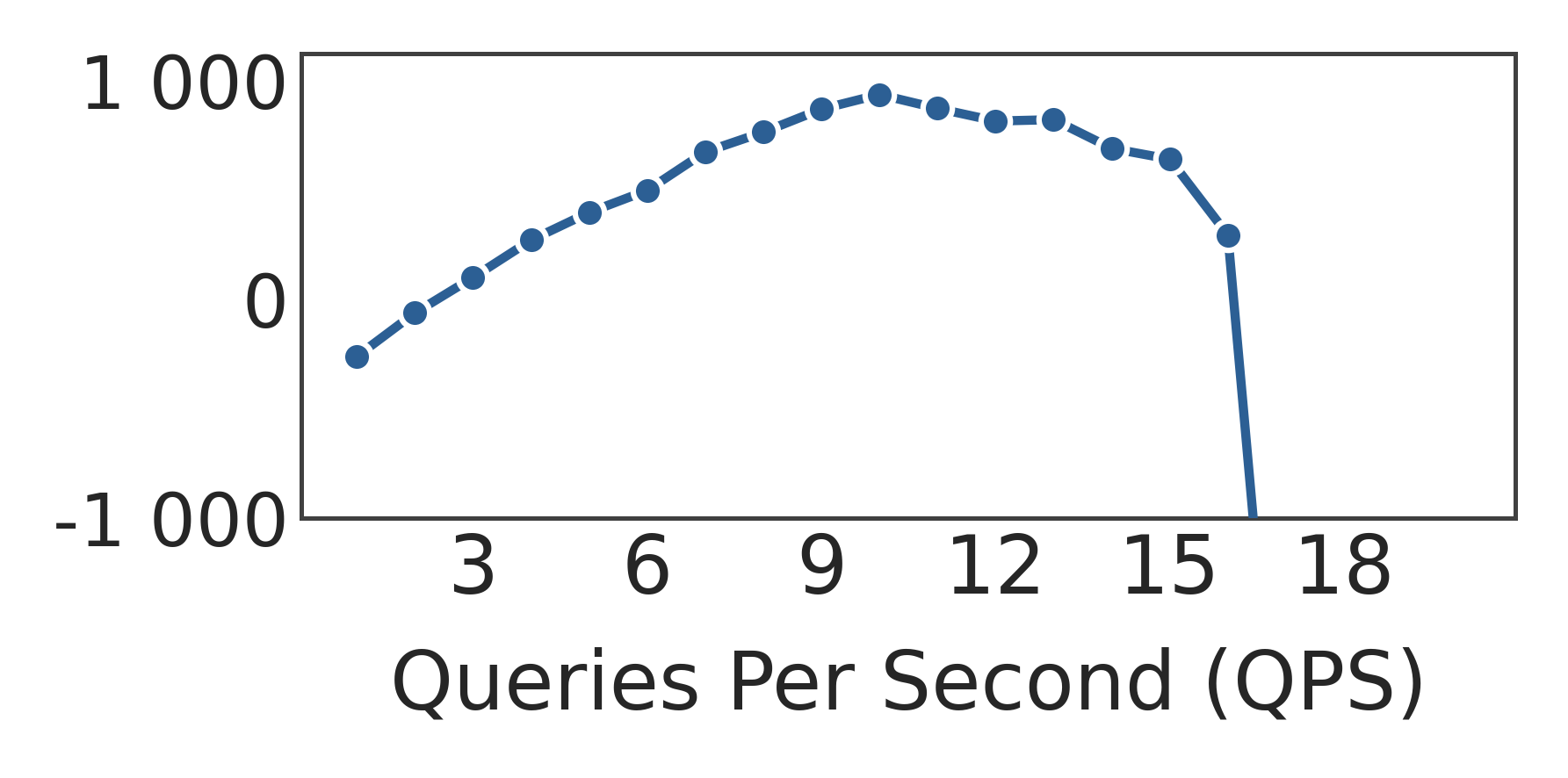}
        \caption{LLaMA-3.1-8B with CP.}
        \label{fig:eval_goodput_llama_chunked}
    \end{subfigure}
    \hfill
    \begin{subfigure}{0.245\textwidth}
        \centering
        \includegraphics[width=0.95\linewidth]{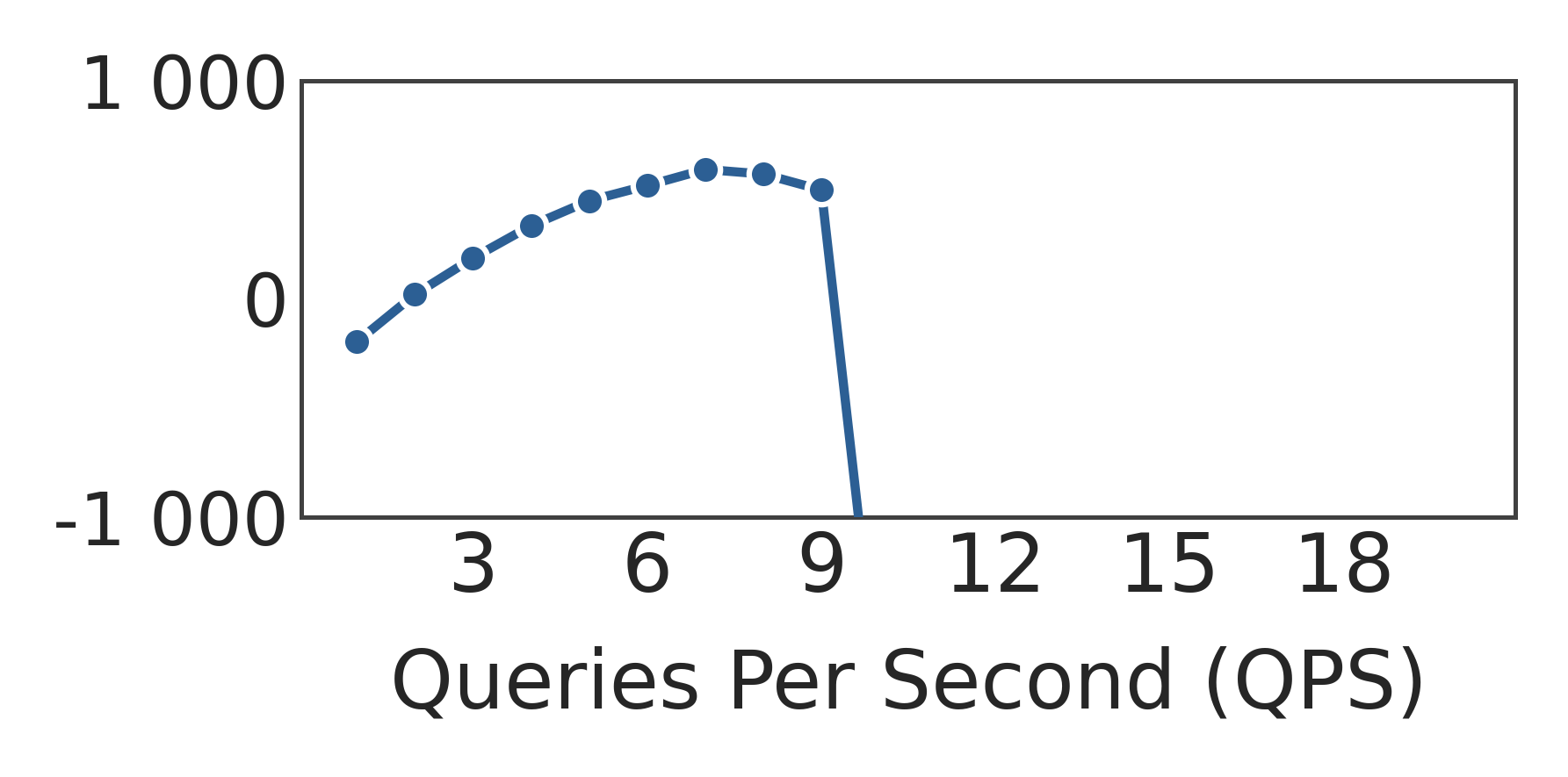}
        \caption{Qwen2.5-14B.}
        \label{fig:eval_goodput_qwen}
    \end{subfigure}
    \hfill
    \begin{subfigure}{0.245\textwidth}
        \centering
        \includegraphics[width=0.95\linewidth]{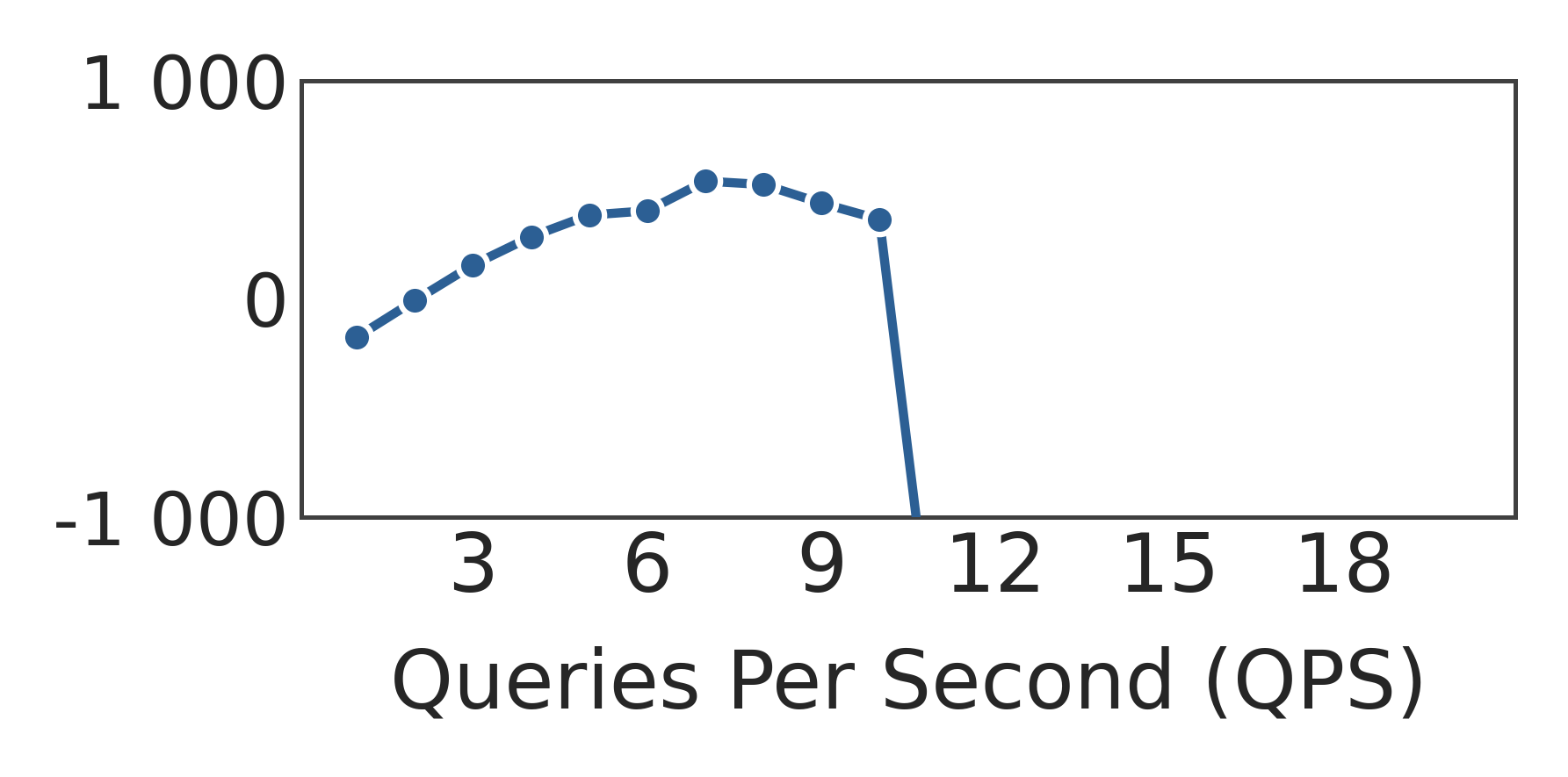}
        \caption{Qwen2.5-14B with CP.}
        \label{fig:eval_goodput_qwen_chunked}
    \end{subfigure}
    \caption{Smooth goodput of vLLM under different QPS, where CP denotes chunked prefills adoption.}
    \label{fig:eval_goodput}
\end{figure*}

The smooth goodput is defined as:
\begin{equation}
    \text{smooth\_goodput} = \frac{\sum_{r\in R} \text{benefit}(r)}{T},
\end{equation}
where $T$ is the time interval of serving the requests committed by the users denoted by $R$.
We notice that Andes~\cite{liu2024andes} also considers the benefit of the requests that miss the SLOs. However, they consider the average token slowdown to the deadline in SLOs, while we consider the maximum token slowdown, i.e., the user idle latency. In practice, once the slowdown has occurred, catching up later does not improve the user experience as the user has already experienced the delay. The maximum slowdown represents the furthest deviation from the deadline within the entire request, which corresponds to the total time the user spends waiting for token generation. Therefore, smooth goodput is more reasonable in this context.

\section{Evaluation}
\label{sec:evaluation}

In this section, we re-evaluate different scheduling strategies under the unified metric framework we propose. Then we analyze the results and summarize the challenges of LLM servings. By comparing with the existing metrics, we demonstrate the advantages of smooth goodput.

\subsection{Experiment Setup}

\stitle{Settings.}
We conduct our experiments on a server equipped with an NVIDIA A100-SXM4-80GB GPU, running Debian GNU/Linux 12 and CUDA 12.2.
We use LLaMA-3.1-8B-instruct~\cite{touvron2023llamaopenefficientfoundation} and Qwen2.5-14B~\cite{qwen2025qwen25technicalreport} as base models in the experiments.
All of our code development is based on vLLM 0.6.3, and the versions of all required packages are consistent with the requirements of it.

\stitle{Workloads.}
For workload, we use ShareGPT~\cite{eccleston2023sharegpt} as the simulation of the conversations with chatbots, and LooGLE~\cite{li2024looglelongcontextlanguagemodels} as the simulation of longer conversations.
We follow~\cite{kwon2023efficient} to set the arrival times of requests to follow the Poisson distribution or processed real-world trace with the average rate set as the parameter to simulate the arrival of requests.
We also conduct the real-world trace experiments to evaluate the performance under real-world scenarios.

\stitle{Metrics.}
We use the smooth goodput to evaluate the performance of LLM serving. As a comparison, we also use the existing SLOs and system level metrics.

\subsection{Analysis with Existing Metrics and Smooth Goodput}
\label{subsec:analysis}

We first analyze the performance of different strategies using existing metrics, highlighting the statistical regularities of vLLM under varying request rates and examining the underlying causes. Subsequently, we introduce smooth goodput under the same scheduling strategy to reveal new insights that existing metrics fail to capture.

\stitle{Analysis with existing metrics.}
Figure~\ref{fig:eval_existing} illustrates the performance of vLLM at different request rates using the ShareGPT dataset, which features relatively short prompts and responses. In the unsaturated stage, as the request rate increases, system resources are utilized more efficiently, leading to a steady increase in throughput thanks to improved parallel processing. At the same time, the increase in batch size results in longer batch processing time and consequently higher values in metrics such as medium TBT and TPOT. When the system reaches its capacity, further increasing the request rate forces more requests into queue, which significantly increases mean TTFT. This analysis reveals that while existing metrics provide a comprehensive overview of service performance, they prioritize throughput and hardware efficiency over the actual user experience. There is no clear indication of a balanced operating point that considers both throughput and the waiting time perceived by users.

\stitle{Analysis with smooth goodput.}
In contrast, smooth goodput quantifies the benefit delivered to the user by taking into account both the throughput and the user idle time. In our experiments, we set the information consumption speed to 5 tokens per second and $\alpha=2.5$. As depicted in Figure~\ref{fig:eval_goodput}, in the unsaturated stage, smooth goodput increases with the request rate because the benefits from additional throughput outweigh the incremental cost of user idle time. However, as the system load continues to rise, the benefit per request starts declining as user idle time becomes the dominating factor, leading to a decrease in smooth goodput. Notably, systems employing chunked prefills reach peak smooth goodput at a higher request rate than those using standard vLLM. This is because the integration of prefill and decode phases through chunked prefills maximizes GPU parallelism, allowing the system to accommodate more requests before queuing delays significantly impact performance. Overall, this expanded analysis underscores the importance of considering both throughput and user experience in LLM serving systems and demonstrates how smooth goodput offers a more balanced and insightful metric than traditional measurements.
\begin{figure}[t]
    \centering
    \begin{subfigure}{0.23\textwidth}
        \centering
        \includegraphics[width=\linewidth]{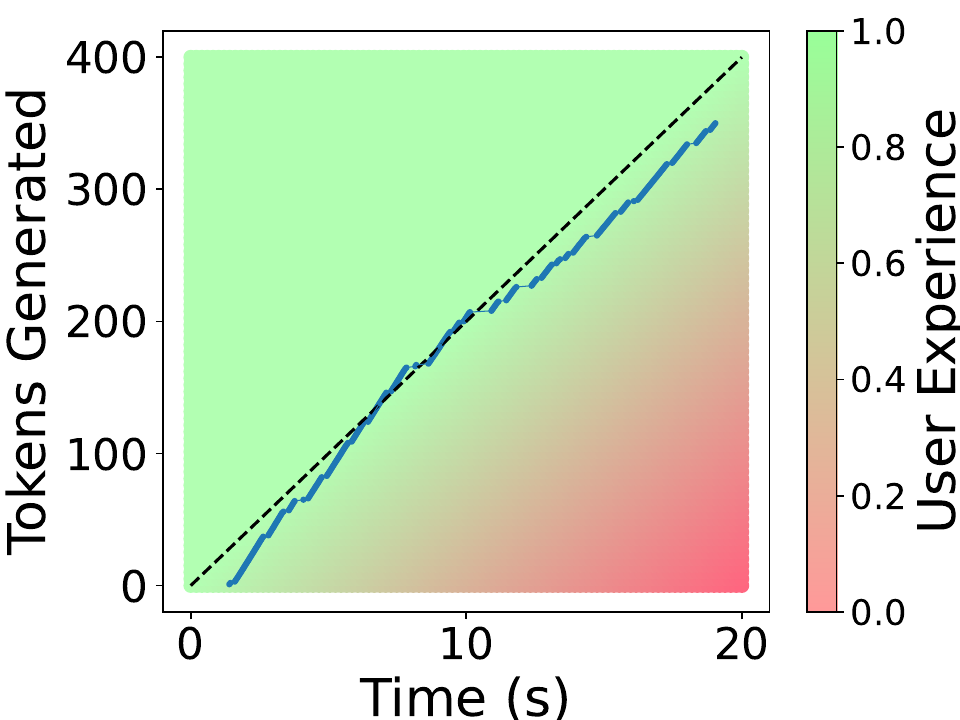}
        \caption{With chunked-prefills off.}
        \label{fig:single_req_vllm}
    \end{subfigure}
    \hfill
    \begin{subfigure}{0.23\textwidth}
        \centering
        \includegraphics[width=\linewidth]{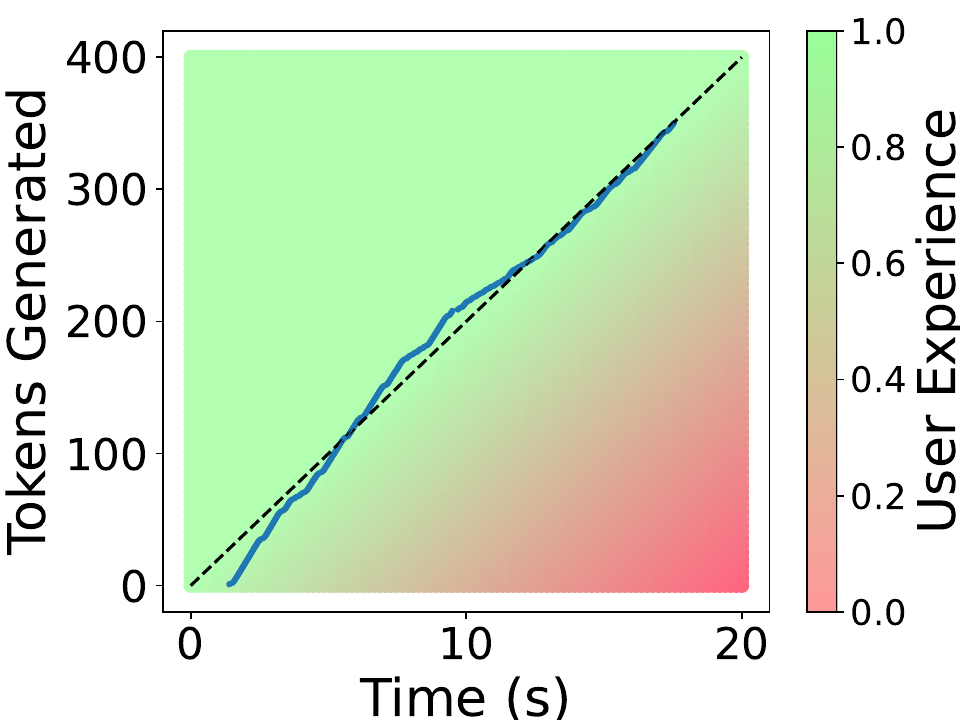}
        \caption{With chunked-prefills on.}
        \label{fig:single_req_sarathi}
    \end{subfigure}
    \caption{Token delivery timeline of vLLM.}
    \label{fig:comparison}
\end{figure}

\begin{figure}[t]
    \centering
    \begin{subfigure}{0.23\textwidth}
        \centering
        \includegraphics[width=\linewidth]{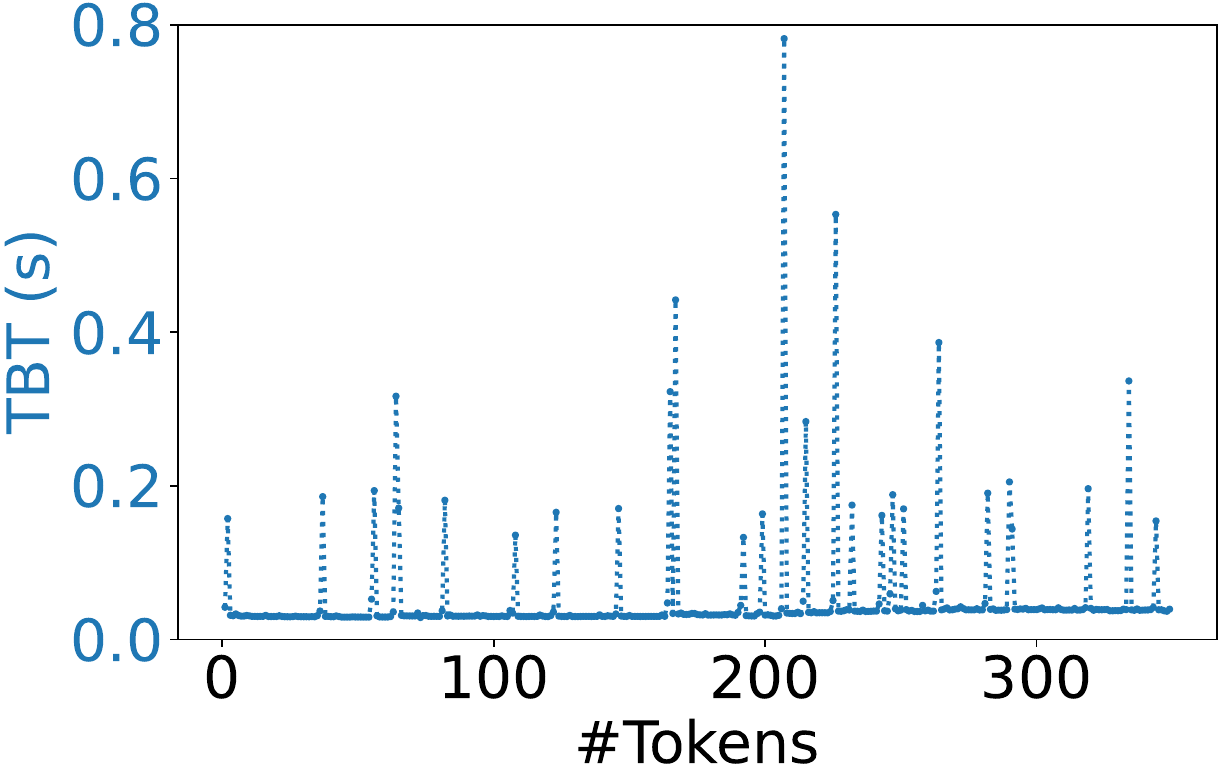}
        \caption{With chunked-prefills off.}
        \label{fig:tbt_vllm}
    \end{subfigure}
    \hfill
    \begin{subfigure}{0.23\textwidth}
        \centering
        \includegraphics[width=\linewidth]{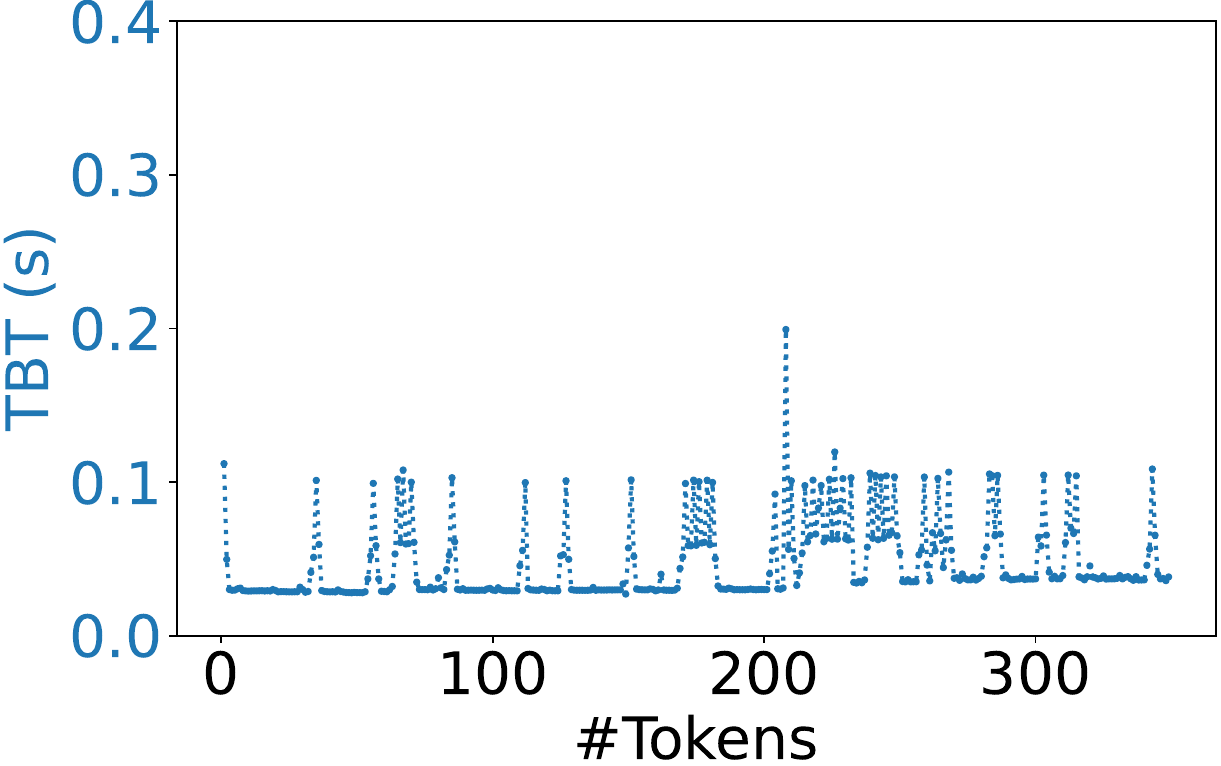}
        \caption{With chunked-prefills on.}
        \label{fig:tbt_sarathi}
    \end{subfigure}
    \caption{TBT of vLLM.}
    \label{fig:tbt}
\end{figure}

\subsection{Analysis with SLOs}

We conduct experiments to systematically demonstrate that our new SLOs can accurately measure the benefit delivered to each request. Our study focuses on prompts averaging around 1600 tokens, and from the service logs of both strategies, we isolate a matching request under identical conditions for a direct comparison.

\stitle{Request level performance analysis.}
Figures~\ref{fig:comparison} and \ref{fig:tbt} illustrate the token generation process for the same request executed with and without the chunked prefills technology. When chunked prefills are applied, the token generation process shows significantly fewer stalls. This improvement stems from the more integrated handling of the prefill and decode stages, resulting in a smoother token delivery timeline. Nonetheless, our data indicates that a number of generation stalls caused by prefill preemption go unnoticed by users—once a sufficient number of tokens have already been delivered, any subsequent stalls have little impact on the perceived responsiveness.

\begin{figure}[t]
    \centering
    \begin{subfigure}{0.235\textwidth}
        \centering
        \includegraphics[width=0.95\linewidth]{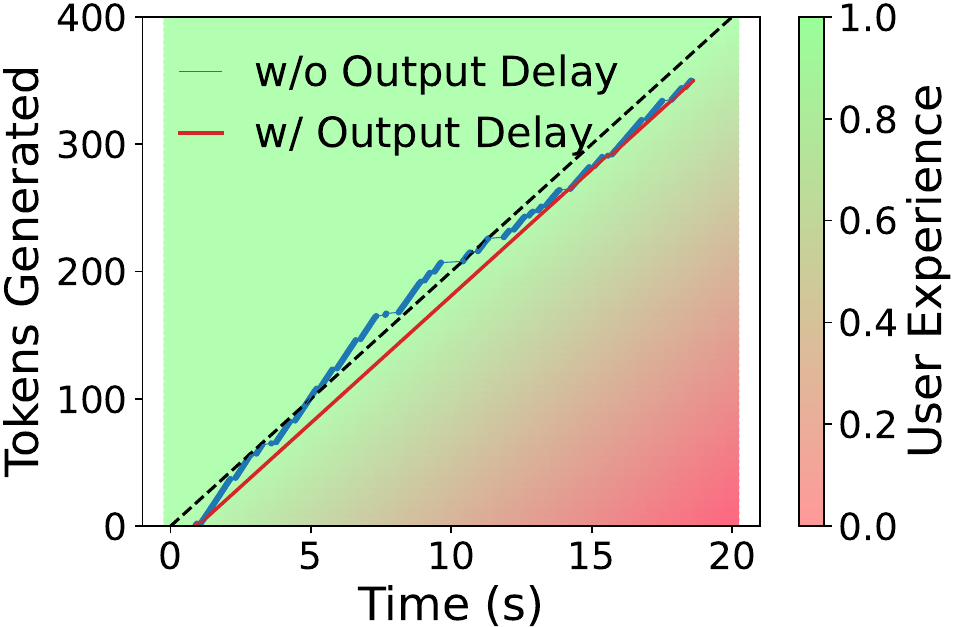}
        \caption{Token delivery timeline.}
        \label{fig:output_delay}
    \end{subfigure}
    \hfill
    \begin{subfigure}{0.235\textwidth}
        \centering
        \includegraphics[width=0.95\linewidth]{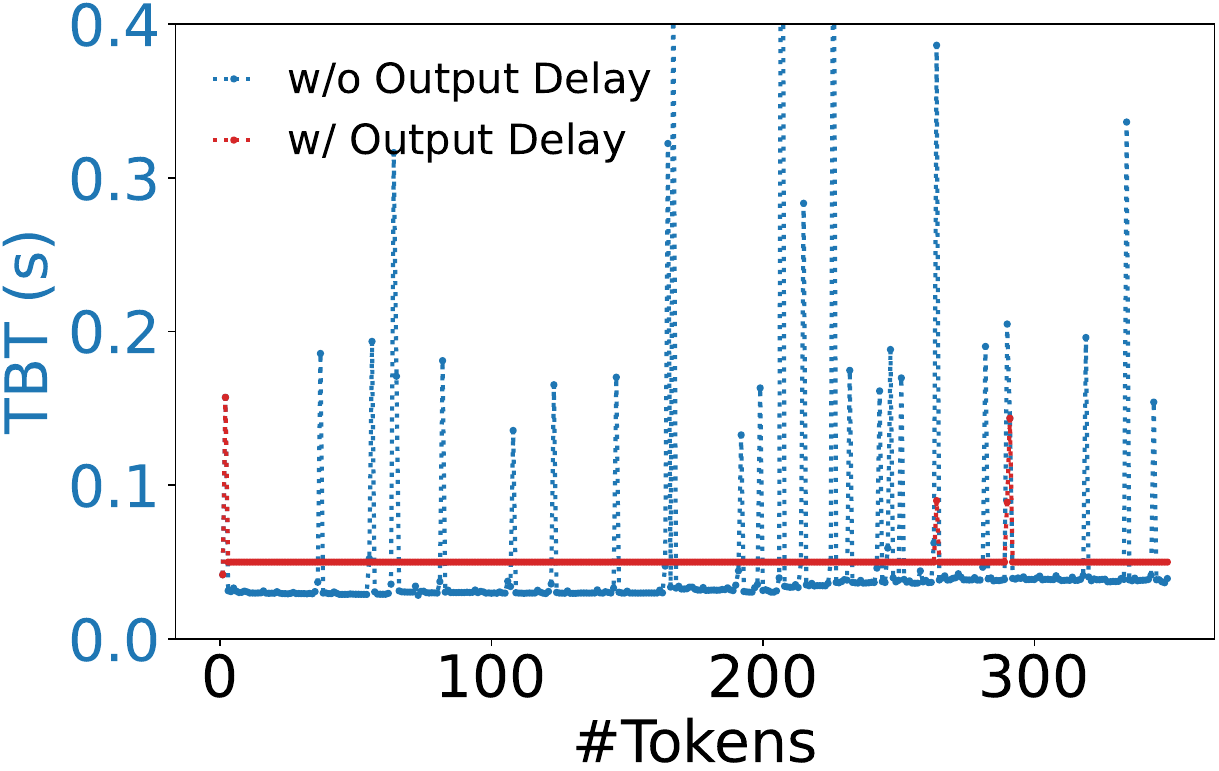}
        \caption{The TBT metric.}
        \label{fig:tbt_delay}
    \end{subfigure}
    \caption{Illustration of the output delay trick.}
    \label{fig:delay}
\end{figure}

\stitle{Output Delaying Trick.}
To further validate our discussion on SLOs, we introduce the output delay trick. As shown in Figure~\ref{fig:output_delay}, this trick works by buffering tokens and releasing them at a deliberately reduced rate. This mechanism is entirely independent of the framework's underlying scheduling strategy, meaning it can be equally implemented on either the server or client side. Compared to the no-delay scenario, the output delay trick effectively reduces the tail TBT, as illustrated by Figure~\ref{fig:tbt_delay}, while leaving overall service throughput unaffected. In essence, although the trick results in a nearly constant TBT—aligning with a user’s typical information consumption rate—it does not reduce the actual user idle time. Consequently, despite better metrics in traditional evaluations, the overall idle time experienced by users remains largely unchanged. This observation reinforces our argument that conventional metrics fall short in capturing true user experience, thereby underscoring the need for more holistic measures like smooth goodput.

\subsection{Extend to Prefill-Decode Disaggregated Architecture}
\label{subsec:disaggregated}
We then show the scalability of our framework for prefill-decode disaggregated LLM serving. For disaggregated structures, the prefill and decode phases are executed on different instances, creating opportunities for decode instances to achieve higher batchsize and less generation stalls.
Therefore, we focus on the decode phase and evaluate the performance with existing metrics and our framework.

Despite for less generation stalls caused by prefill phase, for the requests with long outputs, the lifetime of the decode phase is relatively long, suffering from increased TBT with the length growing. When the request finally violates the SLO, operations such as migration or preemption may be triggered, which can lead to a significant overhead.

\begin{figure}[t]
    \centering
    \includegraphics[width=0.9\linewidth]{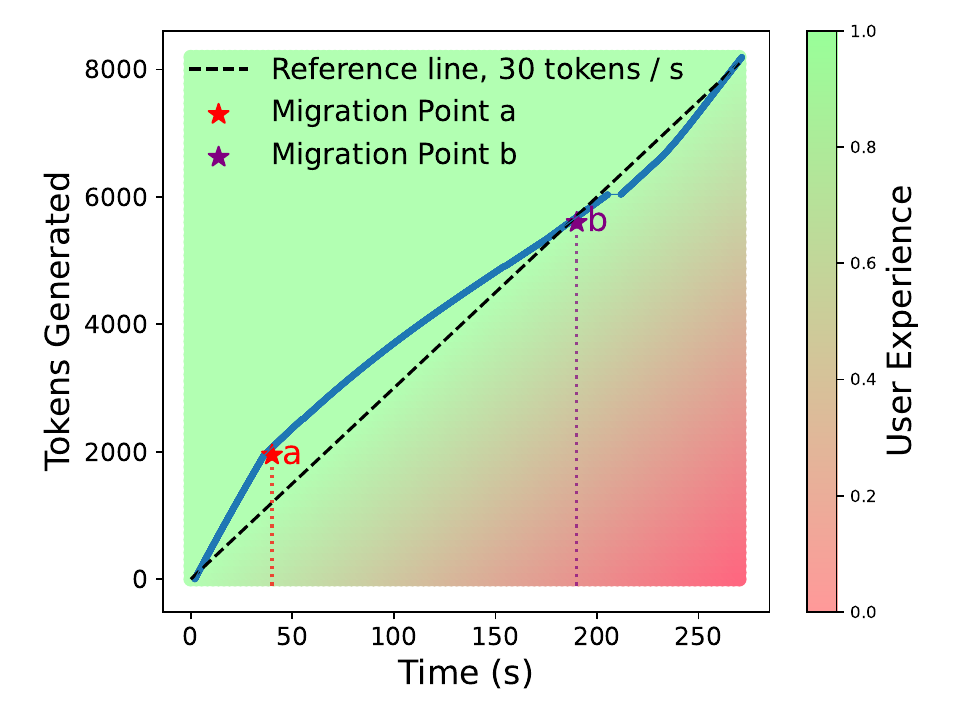}
    \caption{Token delivery timeline in prefill-decode disaggregated architecture.}
    \label{fig:disaggregated}
\end{figure}

Figure~\ref{fig:disaggregated} shows the token delivery timeline of decode phase in disaggregated LLM serving. Since the decode phase is executed on a separate instance, we can observe that the decode phase is not preempted by the prefill phase, resulting in a more stable token delivery timeline. However, it is still affected by the length of the output and batchsize. With the delivery speed decreasing, the tail TBT increases at the red point \textit{a}. However, enough tokens have been delivered at the beginning, so the user experience is not significantly affected in fact. This observation aligns with the smooth goodput, which triggers migration at the purple point \textit{b}, denoting that user will suffer from an idle time without token to read.
Thanks to consideration of the whole timeline, smooth goodput report less migration and preemption events, indicating that the system is able to handle the requests more efficiently without triggering unnecessary operations.

\subsection{Parameter Sensitivity Analysis}

\begin{figure}[t]
    \centering
    \includegraphics[width=0.9\linewidth]{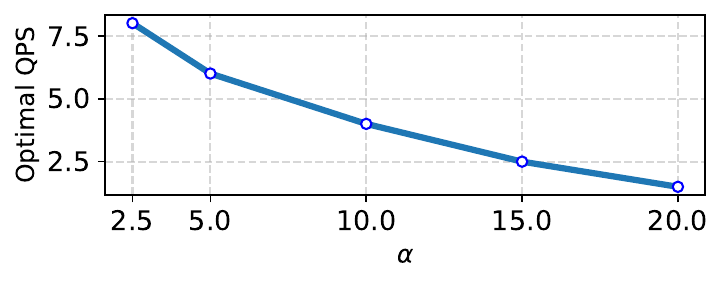}
    \caption{Impact of parameter $\alpha$ on optimal QPS.}
    \label{fig:alpha_impact}
\end{figure}

The parameter $\alpha$ is used to quantify the impact of user idle time on user experience. A larger $\alpha$ corresponds to stricter latency requirements. Different settings of $\alpha$ influence the trend of smooth goodput, thereby affecting the optimal QPS corresponding to the maximum value. Under identical experimental conditions, we configured different $\alpha$ values to demonstrate their impact on the optimal QPS. As shown in Figure~\ref{fig:alpha_impact}, the QPS gradually decreases expectedly as $\alpha$ increases. This is because a larger $\alpha$ places greater emphasis on reducing user idle time, leading to a lower optimal QPS that balances throughput and latency more effectively.

\section{Discussion}
\label{sec:related}

\subsection{Optimizations on Existing Metrics}

In this section, we discuss the system targets on optimizating existing metrics and how smooth goodput can impact their design.

\stitle{Throughput-oriented optimizations.}
Orca~\cite{yu2022orca} introduces the continuous batching, dynamically constructing and processing batches, thereby fully leveraging the parallelism of GPUs.
Building upon this, vLLM~\cite{kwon2023efficient} further incorporates paged attention, which notably enhances compuation throughput, and reduces operational costs.
Consequently, it has been widely adopted and established itself as the SOTA framework for inference services. For these works that prioritize throughput, the SLOs are not explicitly defined. Therefore, we can not directly compare the performance of these works with those that focus on SLO attainment and goodput.

\stitle{SLO attainment-oriented optimizations.}
Splitwise~\cite{patel2023splitwise} and TetriInfer \cite{hu2024inference} proposes splitting prefill and decode phases to separate device due to their different features of computing and memory access.
Sarathi-Serve \cite{agrawal2024taming} introduces chunked prefills and stall-free batching to mitigate the stall of generation.
SCOOT \cite{cheng2025scootsloorientedperformancetuning} propose an automatic paramter tuning system to find the optimal configuration for the system to meet the SLOs.
These works improve SLO attainment defined on different metrics, enabling more requests to be served under SLO requirements.
QLM \cite{patke2024queueneedresolvingheadofline} emphasizes the impact of waiting time and execution time on achieving the SLO on E2E latency. For these works, the SLOs are defined on different metrics, such as TTFT, TBT, and TPOT. Under the same SLO requirements, smooth goodput can be used to compare the performance of these works as a unified framework that combines efficiency and user experience.

\stitle{Goodput-oriented optimizations.}
By avoiding the interference between prefill and decode phases, DistServe~\cite{zhong2024distserve} achieves higher goodput under the same SLO requirements on TTFT and TPOT. That is, more requests that meet the SLOs can be served per second. For the work that designed to improve goodput, smooth goodput attach benefit to all requests in fine-grained level, which is more aligned with user experience.

In summary, despite the diverse metrics, certain projects such as Splitwise, Distserve, and TetriInfer have identified analogous optimization opportunities.
However, the inability to directly compare the effectiveness of these optimizations across different measurement systems poses challenges in making informed optimization choices.

\subsection{Future Work}

For future work, we observe that the latest slow-thinking models~\cite{openai2024openaio1card,deepseekai2025deepseekr1incentivizingreasoningcapability} undergo a lengthy thought process before delivering tokens to users, which highlights the need for more nuanced, semantic-aware SLOs~\cite{shen2025dastdifficultyadaptiveslowthinkinglarge}. This observation motivates us to explore innovative approaches for SLO design where the granularity of SLO requirements can be adjusted according to the semantic content and complexity of the input requests. For instance, we could assign looser SLOs to those requests carrying more intricate or higher-value information, thereby affording the model additional time to process and deliver richer, more detailed responses without compromising overall user satisfaction.

Furthermore, as models evolve in terms of size and reasoning ability, the interplay between output throughput and quality becomes increasingly significant. Future research could investigate adaptive SLO frameworks that dynamically adjust performance requirements based on real-time evaluations of input complexity, system load, and the expected quality of the output. Such a semantic-aware approach would not only balance throughput and user experience more effectively but also pave the way for more resilient and context-sensitive optimization strategies.

In addition, exploring these adaptive SLOs could lead to systems that better accommodate the diverse operational profiles of modern language models, from rapid inference to slow reasoning. This would allow for more refined trade-offs between latency and output quality, ensuring that even when processing demands are higher, the overall user experience remains both consistent and satisfying.

These directions open promising avenues for future work, ultimately aiming to create a unified framework that harmonizes throughput, response quality, and semantic depth in LLM serving.

\section{Conclusion}
In this paper, we propose a metric framework to evaluate the performance of LLM serving. We show that existing metrics fail to capture user experience and demonstrate the correlation between user experience and output delivery speed in streaming LLM serving. We introduce smooth goodput to measure service benefit per unit time, considering both service efficiency and user experience.
Using this framework, we re-evaluate performance under multiple workloads, demonstrating its capability in analyzing service performance. We hope this framework can provide a unified standard for evaluating LLM serving performance and foster research in LLM serving optimization.

\bibliographystyle{ACM-Reference-Format}
\bibliography{reference}

\end{document}